\documentclass[twoside,11pt]{article}

\usepackage{jmlr2e}

\usepackage{amsthm,amssymb}
\usepackage{amsmath}
\usepackage{algorithm2e}
\usepackage{graphicx}
\usepackage{todonotes}
\usepackage{subcaption}
\usepackage{multirow}
\usepackage{tikz}
\usepackage{arydshln}
\usepackage{lineno}

\usepackage{lastpage}
\ShortHeadings{Sparse Linear Centroid-Encoder}{Ghosh and Kirby}
\firstpageno{1}

\begin{document}

\title{Sparse Linear Centroid-Encoder: A Convex Method for Feature Selection}

\author{\name Tomojit Ghosh \email Tomojit.Ghosh@colostate.edu \\
       \addr Department of Mathematics\\
       Colorado State University\\
       Fort Collins, CO 80523 ,USA
       \AND
       \name Michael Kirby \email Kirby@math.colostate.edu \\
       \addr Department of Mathematics\\
       Colorado State University\\
       Fort Collins, CO 80523, USA
       \AND
       \name Karim Karimov \email Karim.Karimov@colostate.edu \\
       \addr Department of Mathematics\\
       Colorado State University\\
       Fort Collins, CO 80523, USA}
\editor{TBD}
\maketitle

\maketitle

\begin{abstract}
We present a novel feature selection technique, Sparse Linear Centroid-Encoder (SLCE). The algorithm uses a linear transformation to reconstruct a point as its class centroid and, at the same time, uses the $\ell_1$-norm penalty to filter out unnecessary features from the input data. The original formulation of the optimization problem is nonconvex, but we propose a two-step approach, where each step is convex. In the first step, we solve the linear Centroid-Encoder, a convex optimization problem over a matrix $A$. In the second step, we only search for a sparse solution over a diagonal matrix $B$ while keeping $A$ fixed. Unlike other linear methods, e.g., Sparse Support Vector Machines and Lasso, Sparse Linear Centroid-Encoder uses a single model for multi-class data. We present an in-depth empirical analysis of the proposed model and show that it promotes sparsity on various data sets, including high-dimensional biological data. Our experimental results show that SLCE has a performance advantage over some state-of-the-art neural network-based feature selection techniques.
\end{abstract}

\section{Introduction}

In the era of Big Data and AI supercomputing, there are unparalled opportunities for knowledge discovery from data.  Deep neural networks and transformers have had a transformative impact on how
scientists, and engineers approach modeling and predictive analytics.  While incredibly powerful, the modeling capabilities of these neural networks are often hard to explain, i.e., they perform well for reasons that are poorly understood. The tendency is to embrace complexity and large parameter sets, rather than simplicity and explainability.  In this paper we present a {\it relatively} simple neural network architecture that has surprisingly good performance even with limited data.

The main modeling question we address here is the existence of a reduced input feature set capable of classifying the phenomenon of interest, and further, if there exists a linear data reduction mapping to facilitate their discovery.
We are motivated along this path given the success nonlinear architectures, e.g.,  deep feature selection (DFS)~\citep{li2016deep} and Centroid Encoder~\citep{ghosh2022supervised}.
For example, the basic idea of Centroid Encoder is to use an autoencoder neural network architecture where the targets at the decoder output are the centroids of the class.  The result is a dimensionality reducing transformation, i.e., the encoder, that can be used for classification.    

Motivated in part by this autoencoder architecture, in this work we propose a linear feature selection tool with a new optimization problem. 
%motivated to simplify success of complex, deep neural networks.
We exploit the linearity of the mapping to create a two step algorithm with each step being convex.  
The two steps are alternated and consist of data fitting
and sparsification.  The major appeal of this approach is the simplicity of the resulting model, the reduced set of explanatory features
and the fact that--given each problem is convex--much less data is actually required
to learn the model.

Why are we modeling {\it small} data sets in the era of Big Data?  While advances in technology have made the acquisition of certain types of health data possible, the ethical issues surrounding the collection of samples from human and animal subjects
make this data extremely limited in comparison to data sets generated by scraping the internet.  Note that in addition to the limited number of available points $p$, the dimension of each sample $n$ can be relatively large, e.g., omics data, so the assumption is that, at least for many biological data sets,  $n \gg p$.

%Can we find other examples of these types of data sets?
%At least describe various omics data sets in more detail.

%{\color{red} We are not allowed to write anything that reveal our identity. It;s a double-blind review. Please rewrite the sentence "We are motivated along this path given our previous work with the nonlinear version of this architecture referred to as Centroid Encoder \cite{ghosh2022supervised}."}

%Discuss motivations for convexity.  Give some optimization theory.

%Add lit review.
Promoting sparsity with an $\ell_1$ penalty for feature selection is now widely used, e.g.,   \citep{tibshirani1996regression,fonti2017feature,muthukrishnan2016lasso,marafino2015efficient,shen2011identifying,lindenbaum2021randomly,candes2008enhancing,daubechies2010iteratively,bertsimas2017trimmed,xie2009scad}. Support Vector Machines \citep{cortes1995support} have also been used extensively for feature selection, see
\citep{marafino2015efficient,shen2011identifying,sokolov2016pathway,guyon2002gene,o2013iterative,chepushtanova2014band}.
%nonlinear models 
%While the linear models are generally fast and convex, they don't capture the non-linear relationship among the input features (unless a kernel trick is applied). %Because of the shallow architecture, these models don't learn a high-level representation of input features. Moreover, there is no natural way to incorporate multi-class data in a single model. 
%Non-linear models based on deep neural networks overcome these limitations. Here, we will briefly discuss a handful of such models. 
Group Sparse ANN  \citep{scardapane2017group} used group Lasso \citep{tibshirani1996regression} to impose the sparsity on a group of variables instead of a single variable. 
%\cite{kim2016opening} proposed a heuristics based technique to assign importance to each feature. Using the ReLU activation, \cite{roy2015feature} provided a way to measure the contribution of an input feature towards hidden activation of next layer. 
An unsupervised feature selection technique based on the autoencoder architecture has been proposed~\citep{han2018autoencoder}. 
%Using a $l_{2,1}$-norm to the weights emanating from each input node, they measure the contribution of each feature while reconstructing the input. The model excavates the input features, which have a minimum contribution. 
Additional examples of nonlinear archtictures include \citep{balin2019concrete,yamada2020feature, singh2020fsnet,taherkhani2018deep}.
 %This algorithm runs the risk of combinatorial explosion for data set with $50K-60K$ features (e.g., microarray gene expression data set).

The contents of this paper is as follows:
In Section \ref{SCLCE} we propose a two step
optimization problem for convex linear feature selection.  In Section \ref{analysis}
we explore the robustness of the methodology including the sensitivity of the method on the sparsity parameter, and the consistency of the features selected.  In Section
\ref{experiments} we do a comparative analysis of the proposed algorithm
with benchmark datasets.  Finally we offer conclusions in Section \ref{dis_cons}.

\section{Sparse Optimization using Linear Centroid-Encoder}
\label{SCLCE}
Consider a data set $X=\{x_i\}_{i=1}^{N}$ with $N$ samples and $M$ classes where $x_i \in \mathbb{R}^d$. Note each sample of data matrix $X$ is represented by a column. The classes are denoted by $C_j, j = 1, \dots, M$ where the indices of the data associated with class $C_j$ are denoted $I_j$. We define centroid of each class as $c_j=\frac{1}{|C_j|}\sum_{i \in I_j} x_i$ where $|C_j|$ is the cardinality of  class $C_j$. Now we define a matrix of class means $\tilde{C} \in \mathbb{R}^{d \times n}$ 
where the $i$'th column of $\tilde{C}$ is the centroid associated
with the class of the $i$'th column of $X$. Note $\tilde{C}$ will have non-unique entries as long as $M<n$. For example, consider the data set $X=\{x_1,x_2,x_3,x_4,x_5\}$ which has two classes $C_1,C_2$ where $I_1=\{1,3,5\}$ and $I_2=\{2,4\}$.  Taking $c_1,c_2$ as the corresponding centroids we have $\tilde C = \{c_1,c_2,c_1,c_2,c_1\}$. With this setup first we will describe Linear Centroid-Encoder (LCE) which is the starting point of our sparse algorithm.

\subsection{Linear Centroid-Encoder (LCE)}
The goal of LCE is to provide the linear transformation of the data
to $k$ dimensions that best approximates the class centroids. Let $A\in \mathbb{R}^{d\times k}$ be the transformation matrix. The unknown matrix ${A}$ may be determined by the following optimization problem
\begin{equation}
\begin{aligned}
\underset {A} {minimize}\;\;\|\tilde C-{A} {A}^T X \|_F^2 \;\;\;
\end{aligned}
\label{equation:LCE_cost}
\end{equation}

Notice that the objective function in Equation (\ref{equation:LCE_cost}) is a convex function of $A$. Let $\mathcal{L}(A)=\frac{1}{2}\|\tilde C-{A} {A}^T X \|_F^2$. The gradient of $\mathcal{L}$ is readily calculated as:

\begin{equation}
\begin{aligned}
\frac{\partial \mathcal{L}}{\partial A}= AA^TXX^TA + XX^TAA^TA - (\tilde C X^T + X \tilde C^T)A
\end{aligned}
\label{equation:LCE_gradient}
\end{equation}

\subsection{Sparse Linear Centroid-Encoder (SLCE)}
Let $B\in \mathbb{R}^{d\times d}$ be a diagonal matrix with $b_{j,j}$ are the diagonal entries. If we left-multiply $X$ by $B$, i.e, $BX$, then each component $x_{i,j}$ of a the sample $x_i$ will be multiplied by $b_{j,j}$ as shown below:\\

$BX = \begin{bmatrix}
    b_{1,1} \;\;0\;\;\hdots \;\; 0\\
    0 \;\; b_{2,2}\;\;\hdots  \;\; 0\\
    %\vdots \;\; \vdots \\
    \vdots \hspace{0.75cm} \vdots  \hspace{0.75cm}\vdots\\
    0 \;\; 0 \;\; \hdots \;\; b_{d,d}\\
\end{bmatrix}
\begin{bmatrix}
    x_{1,1} \;\;x_{2,1}\;\;\hdots \;\;x_{n,1}\\
    x_{1,2} \;\; x_{2,2}\;\;\hdots \;\;x_{n,2}\\
    %\vdots \;\; \vdots \\
    \vdots \hspace{1.0cm} \vdots  \hspace{1.0cm}\vdots\\
    x_{1,d} \;\;x_{2,d} \hdots \;\; x_{d,n}\\
\end{bmatrix} = 
\begin{bmatrix}
    b_{1,1} x_{1,1} \;\;b_{1,1} x_{2,1}\;\;\hdots \;\;b_{1,1} x_{n,1}\\
    b_{2,2} x_{1,2} \;\; b_{2,2} x_{2,2}\;\;\hdots \;\;b_{2,2} x_{n,2}\\
    \vdots \hspace{1.25cm} \vdots  \hspace{1.5cm}\vdots\\
    b_{d,d} x_{1,d} \;\;b_{d,d} x_{2,d} \hdots \;\; b_{d,d} x_{d,n}\\
\end{bmatrix}$

With this setup, we introduce the Sparse Linear Centroid-Encoder below:

\begin{equation}
\begin{aligned}
\underset {A,B} {minimize}\;\;\|\tilde C-{A} {A}^T (BX) \|_F^2 \;\; + \lambda |diag(B)|_1
\end{aligned}
\label{equation:SLCE_cost}
\end{equation}

where $\lambda$ is a hyperparameter. The $\ell_1$-norm will drive most of the $b_{j,j}$ to near zero. As an effect the corresponding elements or features of $x_i$ will be ignored. Hence the the model will work as a linear feature detector. It's noteworthy that our aim is to reconstruct the class centroids $c_j$ of a sample $x_i$ with fewer features. Equation \ref{equation:SLCE_cost} is a non-convex model over the matrices $A,B$. Notice, the optimization becomes convex if we fix one matrix and solve over the other. Hence the solution comes following two convex steps: in the first step we keep $B$ fixed and solve over $A$ and in the second step we freeze $A$ and solve over $B$. The hyperparameter $\lambda$ in Equation \ref{equation:SLCE_conves_step2} controls the sparsity. A higher value will drive most of the $b_{j,j}$ to near zero producing a sparser solution than a solution with lower $\lambda$. Therefore $\lambda$ is the knob which controls the sparsity of our model.

\textbf{Comment of Convexity:} The optimization problem of SLCE is non convex over the matrices $A,B$. But if we optimize over one matrix keeping other one fixed, then the problem becomes convex as shown below:

\begin{equation}
\begin{aligned}
\underset {A} {minimize}\;\;\|\tilde C-{A} {A}^T (BX) \|_F^2 \;\; + \lambda |diag(B)|_1
\end{aligned}
\label{equation:SLCE_conves_step1}
\end{equation}

\begin{equation}
\begin{aligned}
\underset {B} {minimize}\;\;\|\tilde C-{A} {A}^T (BX) \|_F^2 \;\; + \lambda |diag(B)|_1
\end{aligned}
\label{equation:SLCE_conves_step2}
\end{equation}

If we initialize $b_{j,j}$ to 1 of matrix $B$, then Equation (\ref{equation:SLCE_conves_step1}) is equivalent to Equation (\ref{equation:LCE_cost}) which is the LCE cost. First, we solve this optimization which is convex over the set of $\mathbb{R}^ {d \times k}$ matrices $A$. The domain is a convex set, as convex combination of two $\mathbb{R}^ {d \times k}$ matrices is also a $\mathbb{R}^ {d \times k}$ matrix. The cost function $f(A)=\|\tilde C-{A} {A}^T X \|_F^2$ is also convex as any norm is a convex function \citep{10.5555/993483}. The second part of the optimization, i.e., Equation \ref{equation:SLCE_conves_step2} is also convex. The domain is a convex set as any convex combination of two $\mathbb{R}^{d\times d}$ diagonal matrices is also a $\mathbb{R}^{d\times d}$ diagonal matrix. The function $g(B)=\|\tilde C-{A} {A}^T (BX)\|_F^2 \;\; + \lambda |diag(B)|_1 $ is a combination of Frobenius and $\ell_1$ norm which are convex. Hence $g(B)$ is a convex function as it's a combination of two convex functions \citep{10.5555/993483}.

%After that that we keep $A$ fixed and solve the Equation (\ref{equation:SLCE_conves_step2}) over $B$.  

\subsection{Training of SLCE}
As mentioned before, SLCE is a two-step convex algorithm. In the first step, we search for a solution for the matrix $A$ using Equation \ref{equation:LCE_cost} with an embedding dimension set to 5 for all data sets. In this step, we train the model until the absolute value of the difference of the costs of two consecutive iterations becomes less than equal to $10^{-6}$. After that, we fix the matrix $A$ and introduce the diagonal matrix $B$ with diagonal entries set to 1. We run the model for ten iterations without applying the $\ell_1$ penalty to adjust the parameters of $B$. After which, we use the $\ell_1$ penalty on the diagonal elements of $B$ for $2000$ iterations. Throughout the training process, we use a fixed learning rate of 0.002. We don't use mini-batches but train on the entire training set with Adam optimizer \citep{DBLP:journals/corr/KingmaB14}. We implemented SLCE in PyTorch \citep{paszke2017automatic} to run on GPUs on Tesla V100 GPU machines. We will provide the code with a dataset as supplementary material. \\

\textbf{Hyperparameter Tuning:} SLCE uses two hyperparameters: the embedding dimension $k$ and the sparsity parameter $\lambda$. In all of our experiments in the article, we fixed $k=5$. To tune $\lambda$, we did a two-fold cross-validation on a training partition with ten repeats to pick a suitable $\lambda$ for a data set. We chose $\lambda$ from the range $0.04 \dots 0.5$ and the optimal values are kept in Table \ref{table:SLCEHyperparameters_dataset}

\begin{table}[!ht]
	\centering
	\begin{tabular} {|c|c|c|c|c|c|c|}	% (3 columns)
		% start header
		\hline	% Makes 2 fancy lines
		\multirow{2}{*}{Hyperparameter} & \multicolumn{6}{c|}{Dataset} \\
		\cline{2-7}
          & \multicolumn{1}{c|} {ALLAML} & \multicolumn{1}{c|} {GLIOMA} & \multicolumn{1}{c|} {SMK\_CAN} & \multicolumn{1}{c|}{Prostate\_GE} & \multicolumn{1}{c|}{GLI\_85} & \multicolumn{1}{c|}{CLL\_SUB} \\
          \cline{1-7}
          {$\lambda$} & $0.10$ & $0.30$ & $0.05$ & $0.20$ & $0.25$ & $0.04$\\
    \hline
	\end{tabular}	
	\caption{Data set specific sparsity parameter $\lambda$ used in our bench marking experiment.}
	\label{table:SLCEHyperparameters_dataset}
\end{table}

\section{Analysis of SLCE}
\label{analysis}
We did an array of analysis of the proposed model and we present the details here.
\subsection{Feature Sparsity}

\begin{figure}[ht!]
    \centering
    %\vspace{-0.10cm}
    \includegraphics[width=13.75cm, height=4.5cm]{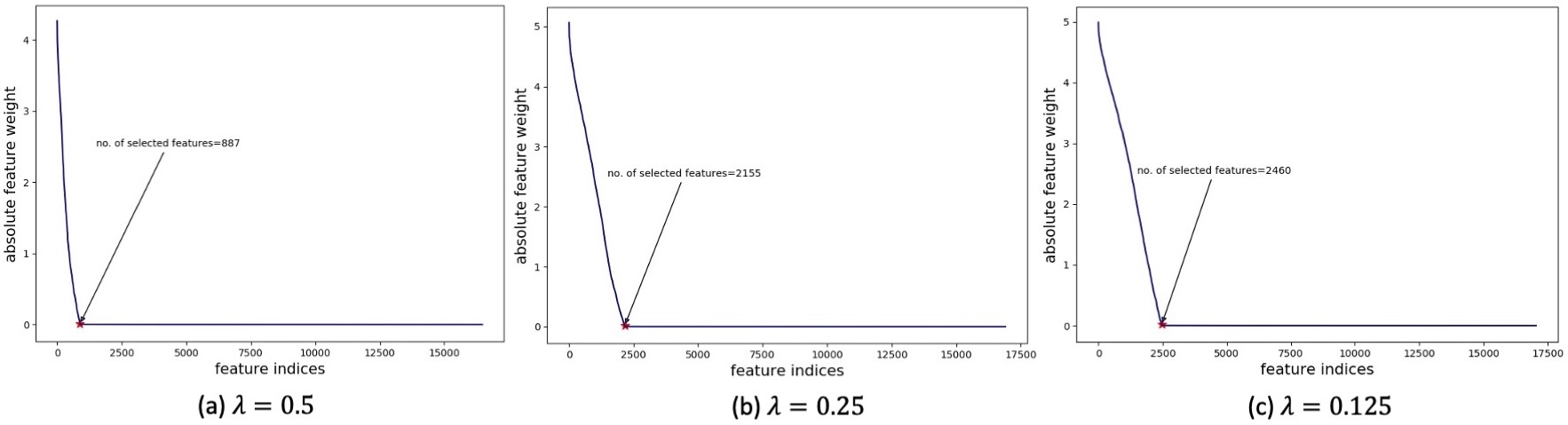}
    %\vspace{-0.3cm}
    \caption{Sparsity analysis of SLCE on Pancan data set for three different choices of $\lambda$. In each case, we plotted the absolute value of weights of the diagonal elements ($b_{jj}$) of matrix $B$ in descending order. These plots suggest the model induces sparsity on the input features, driving most weights to near zero. Notice that as $\lambda$ decreases, the model selects a larger set of features.}
    \label{fig:SLCE_sparsity_analysis}
\end{figure}

The proposed model induces sparsity by minimizing the $\ell_1$-norm of the diagonal matrix $B$. The hyperparameter $\lambda$ acts as a regulator to control feature sparsity; a higher value will generate a sparser solution selecting a small set of features from the input data, whereas a smaller value will pick a large group of variables. Figure \ref{fig:SLCE_sparsity_analysis} shows the sparsity analysis using the high-dimensional Pancan data with $20,531$ features. We split the data set into a $50:50$ ratio of train and test and run SLCE on the training partition. The model promotes feature sparsity by enforcing a significant number of $b_{jj}$ to near zeros ($10^{-4}$ to $10^{-8}$). For example, the model only selects $887$ out of $20,531$ variables of original data when $\lambda=0.5$. As expected, the number of chosen features starts to increase for smaller values of $\lambda$.

%\newpage
\subsection{Analysis of Feature Cut-Off}
\begin{figure}[ht!]
    \centering
    %\vspace{-0.10cm}
    \includegraphics[width=13.75cm, height=4.5cm]{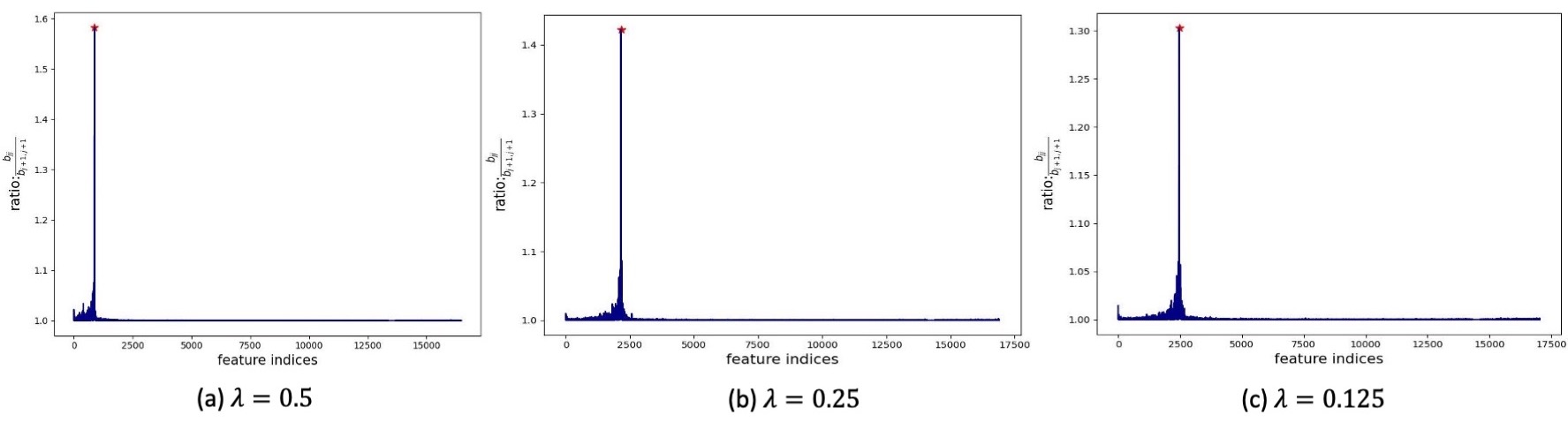}
    %\vspace{-0.3cm}
    \caption{Demonstration of feature cut off using ratio of two consecutive weights.}
    \label{fig:SLCE_feature_cutOff}
\end{figure}

The experiment with feature sparsity shows that the $\ell_1$-norm drives a lot of diagonal elements of matrix $B$ to near zero, which is pretty clear from the three sparsity plots. Here we answer the question of finding out all the features whose absolute weight is significantly higher than the rest. In Figure \ref{fig:SLCE_feature_cutOff}, we plot the ratio of the absolute weight of two consecutive $b_{jj}$'s for three values of $\lambda$. The plot suggests that the ratio is maximum at a specific location. For example, when $\lambda=0.5$, the location is 887, and for $\lambda=0.25$, the location is 2155. Hence one can ignore the features after this position. This location matches the position of the elbow in the sparsity plot in the previous analysis. In all of our experiments, we observe the consistent behavior of the ratio plot, which we use to pick out the significant variables in each run.

\subsection{Feature Selection Stability}
\begin{figure}[ht!]
    \centering
    %\vspace{-0.10cm}
    \includegraphics[width=13.75cm, height=5.5cm]{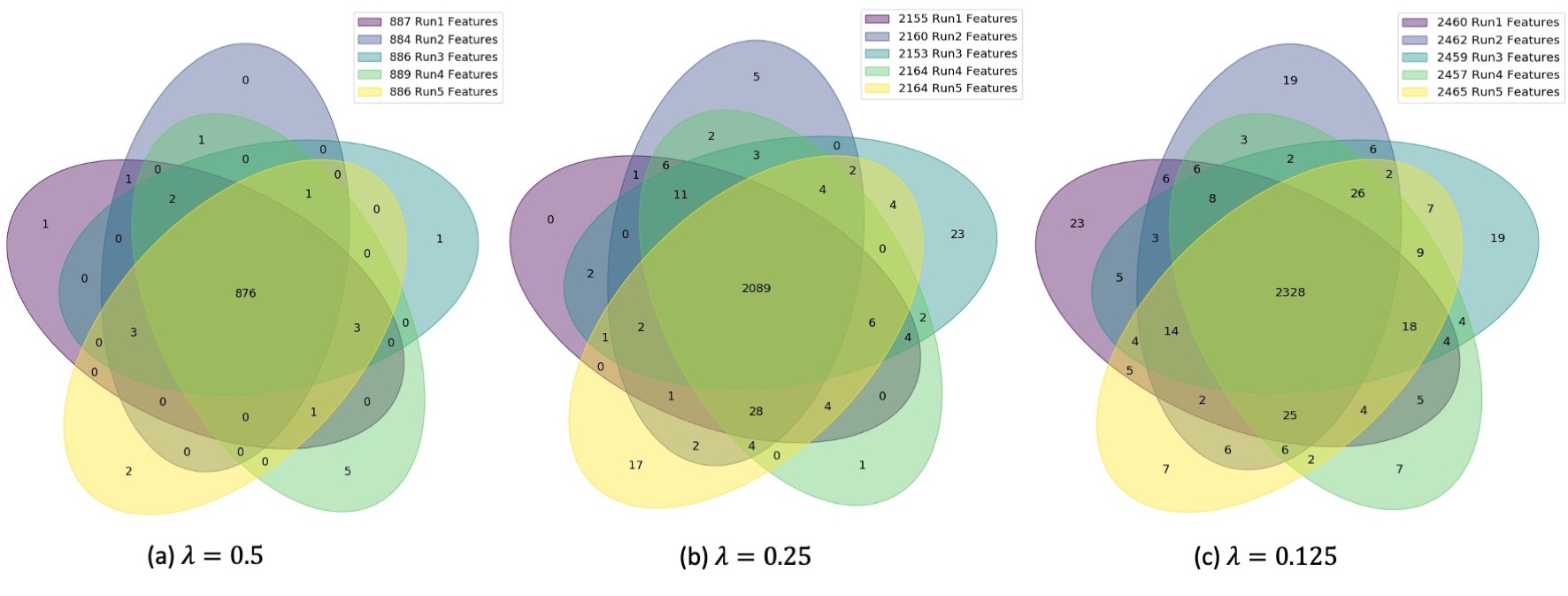}
    %\vspace{-0.3cm}
    \caption{Demonstration of stability of SLCE features on Pancan data set. We ran the model with three choices of $\lambda$ five times each. The Venn diagram shows the intersection of those five feature sets.}
    \label{fig:SLCE_feature_stability}
\end{figure}

Here we analyze how stable the proposed model is regarding the number of selected features over multiple trials. We also check the overlap of the feature sets over several runs to check how similar they are. To this end, we run our model five times on the Pancan data for three different values of $\lambda$ and then compare the feature sets. Figure \ref{fig:SLCE_feature_stability} shows the results. Notice that, in each case, the number of features the model selects is consistent. For example, when $\lambda=0.5$, the model picks 887, 884, 886, 889, and 886 variables with an overlap of 876 resulting a Jaccard similarity of 0.9766. Similarly we found a high Jaccard index of 0.9393 and 0.9006 for $\lambda=$ $0.25$ and $0.125$ respectively. High Jaccard scores indicate that the feature sets have a lot of commonality over different runs.

\subsection{Discriminative Power of SLCE Features}
\begin{figure}[ht!]
    \centering
    %\vspace{-0.10cm}
    \includegraphics[width=13.75cm, height=6.5cm]{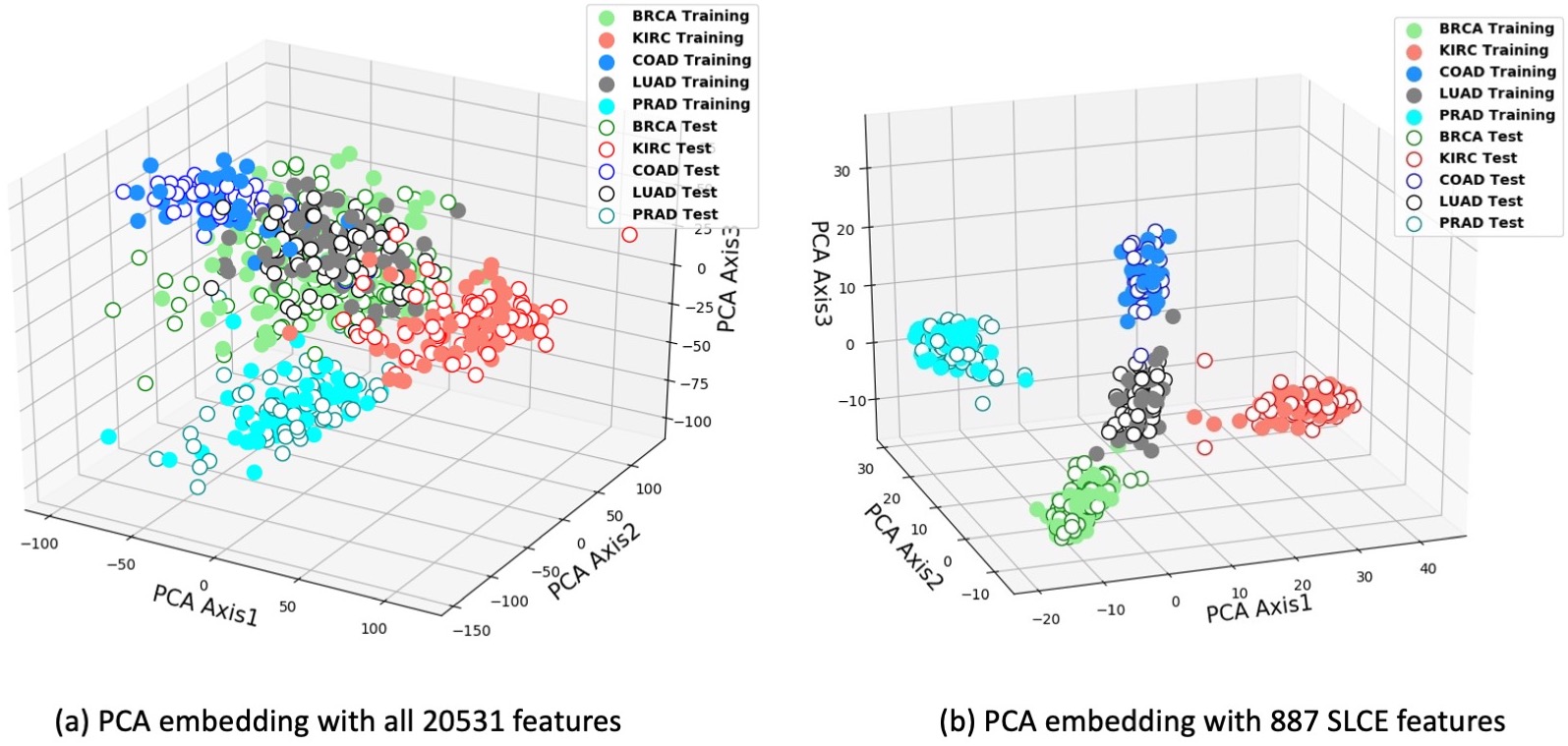}
    %\vspace{-0.3cm}
    \caption{Demonstration of discriminative power of SLCE features on Pancan data set. On the left, we show the three dimensional PCA-embedding done with all the features. On the right, we show the PCA embedding with 887 SLCE features.}
    \label{fig:SLCE_feature_dicriminative_power}
\end{figure}

Now we focus on whether the selected variables help separate the classes, i.e., the discriminative power of selected features. We compare the PCA embedding of Pancan data using all features vs. SLCE-selected features. First, we create the PCA embedding on the training set with all $20,531$ features and show the training and test samples in 3D using a scatter plot. After that, we fit the training set by SLCE with $\lambda=0.5$; pick the selected $887$ features, create a PCA embedding of the training and test samples using the selected features, and compare the 3D scatter plot with the first one
as shown in Figure \ref{fig:SLCE_feature_dicriminative_power}. The embedding with all the features doesn't separate the five classes, whereas PCA does create five distinct blobs of data, one for each tumor type with the SLCE features. Notice that the test samples are also mapped closer to the corresponding training data.

\newpage
\section{Experimental Results}
\label{experiments}
We present the comparative evaluation of our model on various data sets using several feature selection techniques. 

\subsection{Experimental Details}

\begin{table}[!ht]	
	%\vspace{-4mm}
	\centering
	\begin{tabular} {|c|c|c|c|c|c|}	% (3 columns)
		% start header
		\hline	% Makes 2 fancy lines
		Dataset & No. Features & No. of Classes & No. of Samples & Domain \\
		\hline
		ALLAML & 7129 & 2 & 72 & Biology \\
		GLIOMA & 4434 & 4 & 50 & Biology \\
		SMK\_CAN & 19993 & 2 & 187 & Biology \\
		Prostate\_GE & 5966 & 2 & 102 & Biology \\
		GLI\_85 & 22283 & 2 & 85 & Biology \\
		CLL\_SUB & 11340 & 3 & 111 & Biology \\
		PanCan & 20531 & 5 & 801 & Biology(RNA-Seq) \\
		\hline
	\end{tabular}	
 \medskip
	\caption{Descriptions of the data sets used for benchmarking experiments.}
	\label{table:dataDescription}
\end{table}

We use seven high-dimensional biological (see  
Table \ref{table:dataDescription}) to compare SLCE with Penalized Fisher's Linear Discriminant Analysis (PFLDA) \citep{witten2011penalized} and three neural network-based models to run benchmarking experiments. We implemented PFLDA in Python to compare and contrast with SLCE. Apart from comparing classification result, we also check the sensitivity of sparsity parameter of these two linear models. For benchmarking with ANN-based methods, we picked the published results from the article \cite{singh2020fsnet}, except for Stochastic Gates \citep{yamada2020feature}, which we ran by ourselves using authors code from GitHub. We followed the same experimental methodology described in \citep{singh2020fsnet} for an apples-to-apples comparison. This approach permitted a direct comparison of FsNet, Supervised CAE using the authors' best results. The experiment follows the following workflow:

\begin{itemize}
    \item Split each data sets into training and test partition using 50:50 ratio.
    \item Run SLCE on the training set to extract top $K\in\{10,50\}$ features.
    \item Using the top $K$ features train a one hidden layer ANN classifier with $500$ ReLU units to predict the test samples.
    \item Repeat the classification 20 times and report average accuracy.
\end{itemize}

\subsection{Result: SLCE vs  PFLDA}
\begin{figure}[ht!]
    \centering
    %\vspace{-0.10cm}
    \includegraphics[width=8.00cm, height=5.5cm]{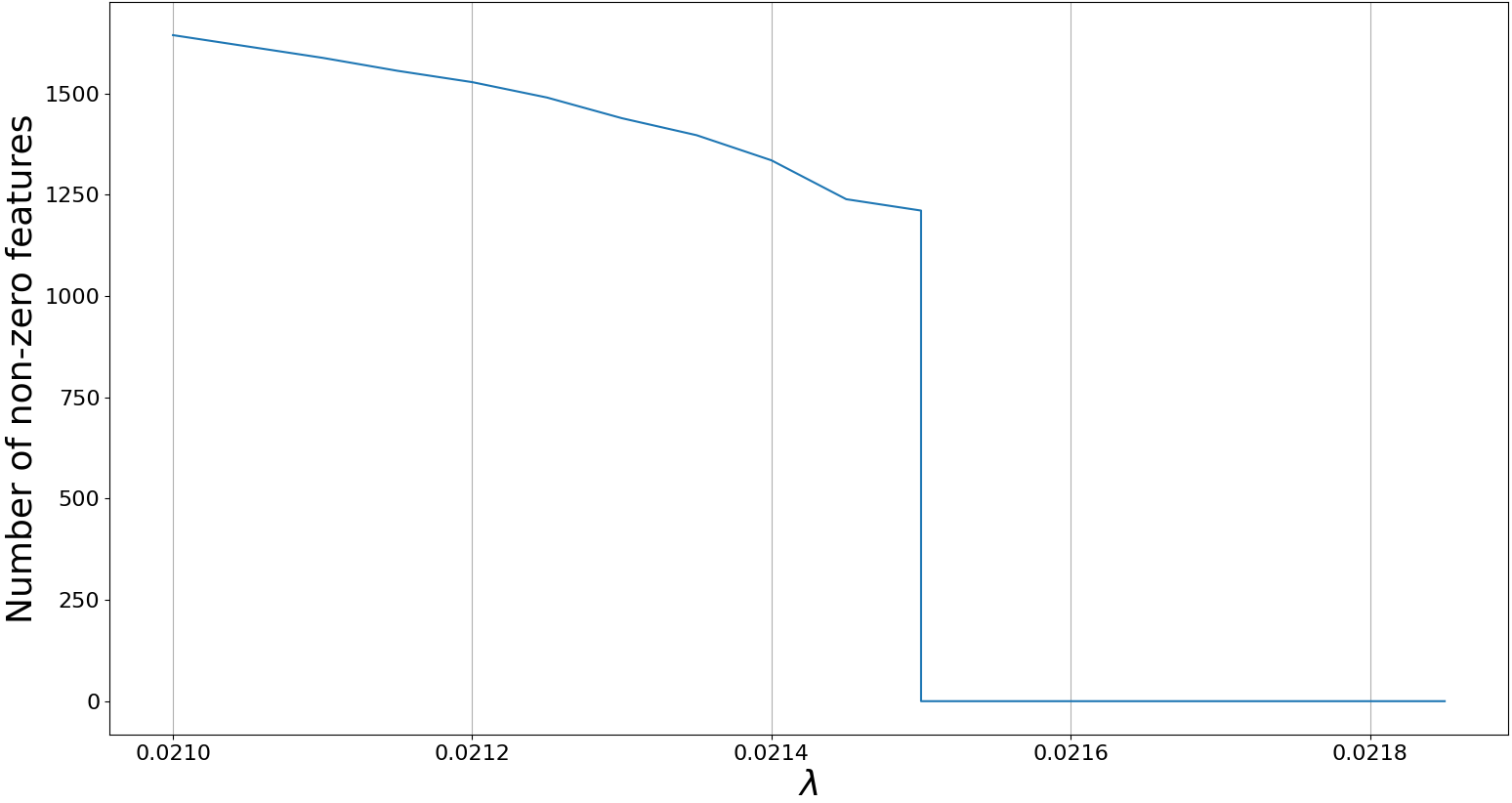}
    %\vspace{-0.3cm}
    \caption{Sparsity analysis of PFLDA on ALLAML data using different values of $\lambda$.}
    \label{fig:SLDA_Sparsity_ALLAML}
\end{figure}

\begin{figure}[ht!]
    \centering
    %\vspace{-0.10cm}
    \includegraphics[width=8.0cm, height=5.0cm]{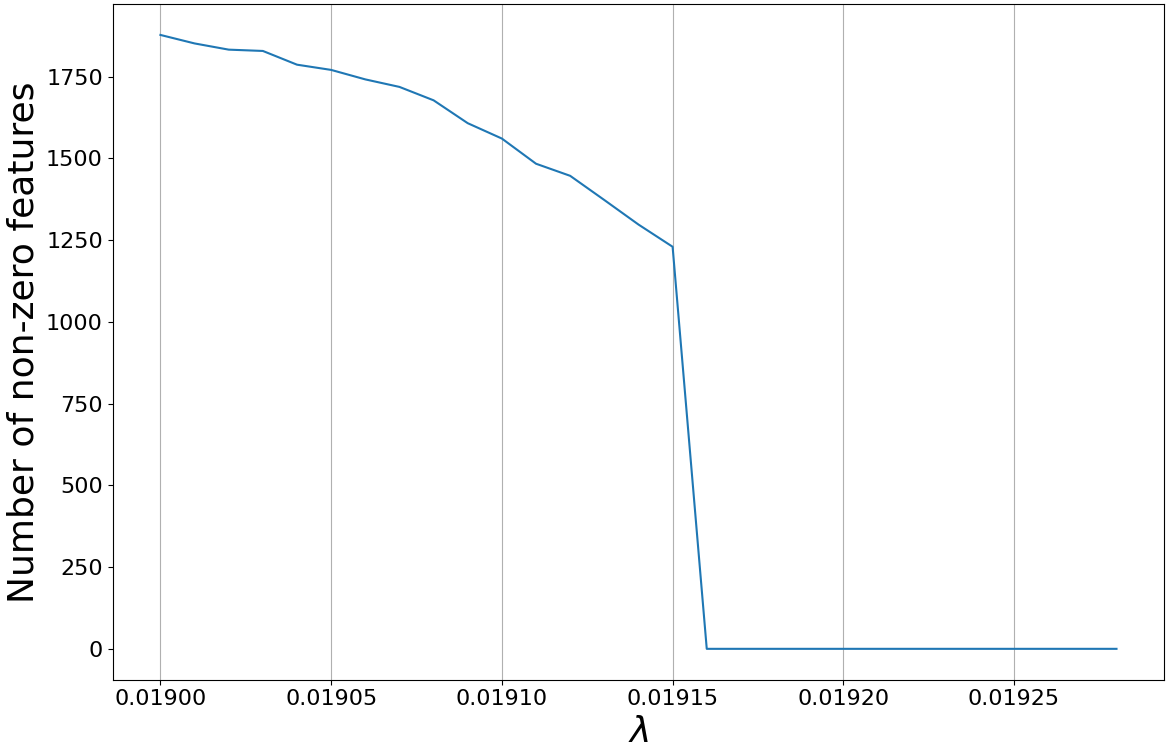}
    %\vspace{-0.3cm}
    \caption{Sparsity analysis of PFLDA on Prostate\_GE data using different values of $\lambda$.}
    \label{fig:SLDA_Sparsity}
\end{figure}

Table \ref{table:exp1_results} compares classification accuracies using the top features of PFLDA and SLCE. We should note that in our experiments we found PFLDA to be very sensitive to the choice of $\lambda$, which controls sparsity. For example, setting $\lambda=0.021$ works for the model on ALLAML data, but appears to behave spuriously for a value of 0.0215 returning no non-zero features, see Figure \ref{fig:SLDA_Sparsity_ALLAML}. We also found that the PFLDA model's performance depends significantly on the data dimension becoming increasingly sensitive as the ambient dimension grows. For example, we observed that the PFLDA model abruptly sets the values of the weights of all the features to zero if $\lambda$ exceeds a threhold as shown in Figure \ref{fig:SLDA_Sparsity}; in other words, no features are selected. 
%Notice that the model by design sets the weight of all the features to zero when $\lambda$ is increased slightly. As an effect the model doesn't return a feature set.

\begin{table}[ht!]
	%\vspace{-0.25cm}
	\centering
	\begin{tabular} {|c|c|c|c|c|c|}	
		% start header
		\hline 	% Makes 2 fancy lines
		%\multicolumn{1}{|c|}{Method} & \multicolumn{3}{c|} {Dataset}\\
		\multirow{2}{*}{Data set} & \multicolumn{2}{c|} {Top 10 features} & \multicolumn{2}{c|} {Top 50 features}  & \multirow{1}{*}{All Features} \\
		\cline{2-5}
		& \multicolumn{1}{c|} {PFLDA} &  \multicolumn{1}{c|} {SLCE} & \multicolumn{1}{c|} {PFLDA} & \multicolumn{1}{c|} {SLCE} & \multirow{1}{*}{ANN} \\
		
		\hline
		ALLAML & $92.7$ & $\textbf{94.1}$ & $95.8$ & $\textbf{96.1}$ & $89.9$\\
        \hline
        Prostate\_GE & $90.7$ & $\textbf{91.2}$ & $88.3$ & $\textbf{90.7}$ & $75.9$\\
		\hline
        GLIOMA & $\textbf{59.9}$ & $58.8$ & $66.2$ & $\textbf{69.2}$ & $70.3$\\
		\hline
		SMK\_CAN & $66.6$ & $\textbf{67.3}$ & $68.6$ & $\textbf{70.9}$ & $65.7$\\
		\hline
        GLI\_85 & $\textbf{85.8}$ & $84.9$ & $85.3$ & $\textbf{85.5}$ & $79.5$\\
        \hline
        CLL\_SUB & $52.5$ & $\textbf{62.9}$ & $52.3$ & $\textbf{75.3}$ & $56.9$\\
        \hline
		
	\end{tabular}
 \medskip
	\caption{Comparison of mean classification accuracy of PFLDA, and SLCE features on six real-world high-dimensional biological data sets. The prediction rates are averaged over twenty runs on the test set. The result on the last columns is computed using all the features from one hidden layer neural network with 500 ReLU units. }
	\label{table:exp1_results}
\end{table}

Due to this limitation of PFLDA, the model took significant time to find a suitable $\lambda$. Returning to the comparison in Table \ref{table:exp1_results}, we see that generally, the SLCE features produce better classification performance than PFLDA. The top 50 SLCE features more accurately predict the test samples than the top 50 PFLDA features in all the cases. In the 'Top 10 features' group, SLCE performed better in four out of six cases. These results establish the performance advantage of our proposed method over PFLDA. The last column of the table shows the accuracy with all the features using a single hidden layer neural network classifier with 500 ReLU units. Notice that the classification using top 50 SLCE features is better than all features, except for GLIOMA. 

\subsection{Result: SLCE vs ANN-Based Models}
\begin{figure}[ht!]
    \centering
    %\vspace{-0.10cm}
    \includegraphics[width=13.75cm, height=8.5cm]{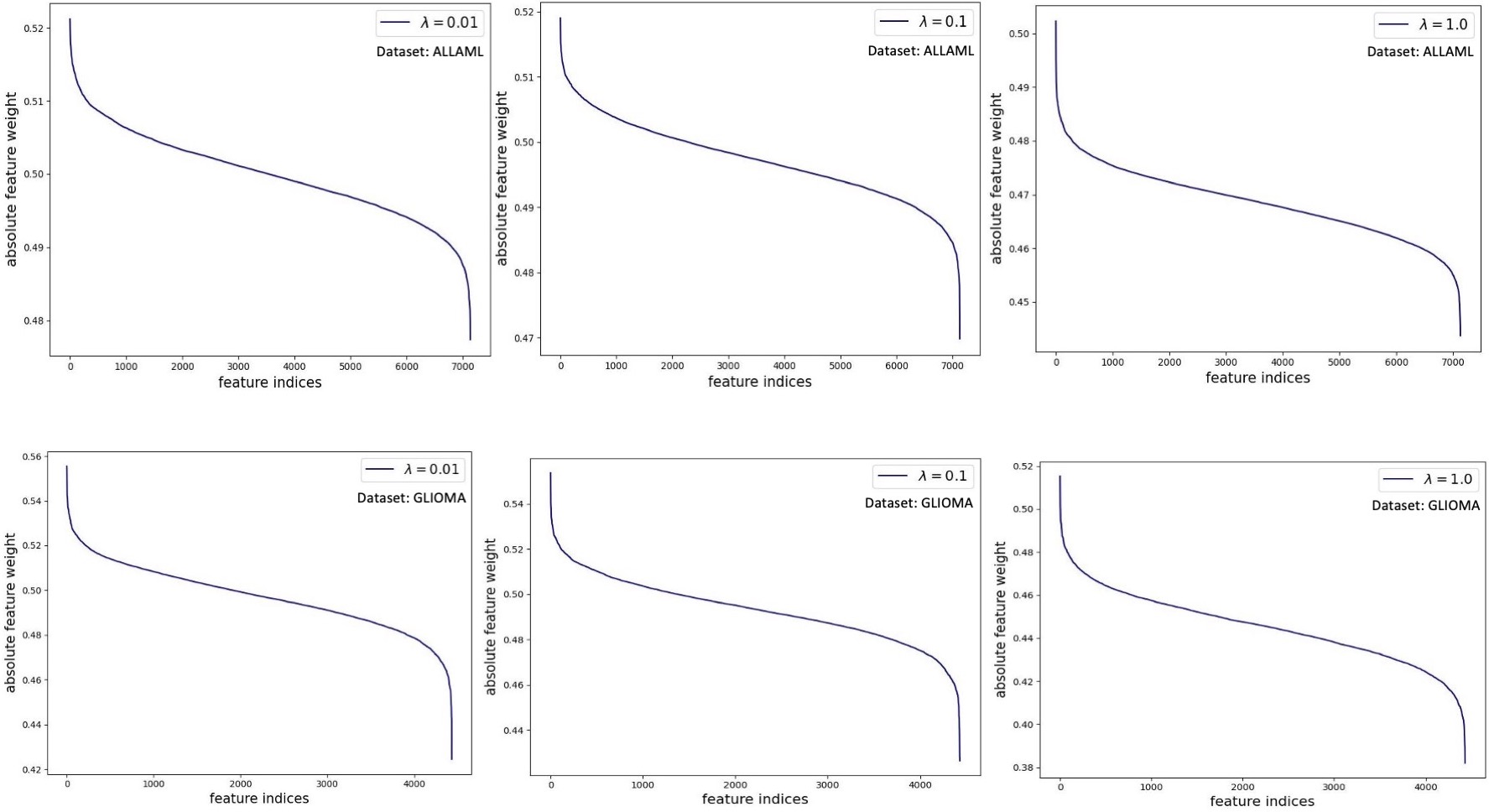}
    %\vspace{-0.3cm}
    \caption{Sparsity analysis of Stochastic Gates on ALLAML and GLIOMA data using three values of $\lambda$.}
    \label{fig:STG_Sparsity}
\end{figure}

Table \ref{table:exp2_results} presents the classification performance using the top features of Feature Selection Network (FsNet), Supervised Concrete Autoencoder (SCAE), Stochastic Gate (STG), and Sparse Linear Centroid-Encoder (SLCE). Note that apart from SLCE, the other methods are neural network-based nonlinear models. We also present the classification accuracy for each data set using all the features under the column "All Fea." Generally, feature selection helps to improve classification accuracies. From the category of top ten features, FsNet produces the best result in three cases, followed by SLCE, with the best accuracy in two cases. SCAE surpasses the other two models in GLI\_85 data. Notice Stochastic Gates doesn't perform competitively in this category. On the other hand, STG is the top-performing model in one case in the top 50 features category, where SLCE outperforms the other models in the remaining five data sets. Notice that the performance of FsNet doesn't improve with more features (e.g., GLIOMA); in fact, the accuracies drop in CLL\_SUB, GLI\_85, SMK\_CAN. We observe the same trend for SCAE in those three data sets. In contrast, SLCE and STG benefit from more features. Considering all the twelve classification tasks, our proposed method performed the best in seven cases, followed by FsNet (three best results).
\begin{table}[h!]
	%\vspace{-0.25cm}
	\centering
	\begin{tabular} {|c|c|c|c|c|c|c|c|c|c|}	% (3 columns)
		% start header
		\hline 	% Makes 2 fancy lines
		%\multicolumn{1}{|c|}{Method} & \multicolumn{3}{c|} {Dataset}\\
		\multirow{2}{*}{Data set} & \multicolumn{4}{c|} {Top 10 features} & \multicolumn{4}{c|} {Top 50 features}  & \multirow{1}{*}{All} \\
		\cline{2-9}		
		& \multicolumn{1}{c|} {FsNet} &  \multicolumn{1}{c|} {SCAE} & \multicolumn{1}{c|} {STG} & \multicolumn{1}{c|} {SLCE} & \multicolumn{1}{c|} {FsNet} & \multicolumn{1}{c|} {SCAE} & \multicolumn{1}{c|} {STG} & \multicolumn{1}{c|} {SLCE} & \multirow{1}{*}{Fea.} \\
		
		\hline
		ALLAML & $91.1$ & $83.3$ & $81.0$ & $\textbf{94.1}$ & $92.2$ & $93.6$ & $88.5$ & $\textbf{96.1}$ & $89.9$\\
        \hline
        Prostate\_GE & $87.1$ & $83.5$ & $82.3$ & $\textbf{91.2}$ & $87.8$ & $88.4$ & $85.0$ & $\textbf{90.7}$ & $75.9$\\
		\hline
        GLIOMA & $\textbf{62.4}$ & $58.4$ & $62.0$ & $58.8$ & $62.4$ & $60.4$ & $\textbf{70.4}$ & $69.2$ & $70.3$\\
		\hline
		SMK\_CAN & $\textbf{69.5}$ & $68.0$ & $65.2$ & $63.1$ & $64.1$ & $66.7$ & $68.0$ & $\textbf{70.9}$ & $65.7$\\
		\hline
        GLI\_85 & $87.4$ & $\textbf{88.4}$ & $72.2$ & $84.9$ & $79.5$ & $82.2$ & $81.0$ & $\textbf{85.5}$ & $79.5$\\
        \hline
        CLL\_SUB & $\textbf{64.0}$ & $57.5$ & $54.4$ & $62.9$ & $58.2$ & $55.6$ & $63.2$ & $\textbf{75.3}$ & $56.9$\\
        \hline
		
	\end{tabular}
 \medskip
	\caption{Comparison of mean classification accuracy of FsNet, SCAE, STG, and SLCE features on six real-world high-dimensional biological data sets. The prediction rates are averaged over twenty runs on the test set. Numbers for FsNet and SCAE are being reported from \citep{singh2020fsnet}. The result on the last columns is computed using all the features from one hidden layer neural network with 500 ReLU units.}
	\label{table:exp2_results}
\end{table}
We analyzed the STG's ability to promote feature sparsity using ALLAML and GLIOMA. We took three values of $\lambda$ to be 0.01, 0.1 and 1.0 to fit the model on a training partition. After that we plotted the probability of the gates in descending order. Figure \ref{fig:STG_Sparsity} presents the sparsity plot. Notice the model failed to promote feature sparsity on these two data sets. Er observed the similar pattern for other data sets as well. 

\section{Discussion, Conclusion and Limitations}
\label{dis_cons}
In this research work, we proposed a novel feature selection model, Sparse Linear Centroid-Encoder (SLCE). We presented a convex training approach for SLCE in two steps. Being linear, it is well suited for low sample size and high-dimensional biological data sets. The model doesn't require an exhaustive search of network architecture and other training-related hyperparameters, typical to a neural network-based model. The convex training approach guarantees that each step has a global solution that is advantageous compared to nonconvex methods. Unlike other linear and non-linear methods, e.g., Lasso, and Sparse SVM, the model doesn't use the response variable in data fitting; instead, SLCE finds a linear transformation to reconstruct a sample as its class centroid followed by a sparse training using $\ell_1$-norm. These innovations make our model unique.

SLCE also enjoys the benefits of a single model for multiclass data—the feature selection mechanism using a diagonal matrix work globally over all the classes on the data set. This aspect of SLCE makes it attractive over other linear techniques, e.g., Lasso and SSVM, where a binary feature selection method is used as a multiclass method by one-against-one(OAO) or one-against-all (OAA) class pairs. These models will suffer a combinatorial explosion when the number of classes increases. Our model also has a significant benefit over Penalized Fisher's Discriminant Analysis for multiclass problems. PFLDA induces sparsity in all directions making it hard to find a single set of features necessary for all the FLDA directions. In contrast, our proposed method searches for a global group of features in the feature selection stage. Unlike PFLDA, our model is not sensitive to the sparsity-inducing parameter $\lambda$. 

We showed that the model successfully enforces sparsity on numerous biological data sets using $\ell_1$-norm. The sparsity parameter controls the number of selected features as expected. The visualization experiment with the Pancan data shows how the model select discriminative feature that improves the PCA embedding. We also found that the variables chosen over multiple runs have many similarities regarding feature count and feature overlap. The visualization with the Venn diagram and the Jaccard index supports the claim.

The extensive benchmarking with six biological data sets and five methods provides evidence that the features of SLCE often produce better generalization performance than other state-of-the-art models. Apart from the linear method PFLDA, we compared SLCE with four ANN-based state-of-the-art feature selection techniques and found that it produced the best result in six cases out of twelve classification experiments. Unlike Stochastic Gates, we have found that our model consistently sparsifies the input features for biological datasets. The strong generalization performance, coupled with the ability to sparsify input features, establishes the value of our model as a linear feature detector.

SLCE, in its current form, maps a sample to its class centroid while applying sparsity. The features may not be discriminatory if two class centroids are close in the ambient space. Adopting a cost that also caters to separating the classes may be beneficial. Our model may not be the right choice for the cases where class centroids make little sense, e.g., natural images. The current scope of the work doesn't allow us to investigate other optimization techniques, e.g., proximal gradient descent, trimmed Lasso, etc., which we plan to explore in the future.

%\section*{References}
\bibliography{SLCE}

%%%%%%%%%%%%%%%%%%%%%%%%%%%%%%%%%%%%%%%%%%%%%%%%%%%%%%%%%%%%
\end{document}